\documentclass[conference]{IEEEtran}
\IEEEoverridecommandlockouts
\usepackage{cite}
\usepackage{amsmath,amssymb,amsfonts}
\usepackage{algorithmic}
\usepackage{graphicx}
\usepackage{textcomp}
\usepackage{xcolor}
\def\BibTeX{{\rm B\kern-.05em{\sc i\kern-.025em b}\kern-.08em
    T\kern-.1667em\lower.7ex\hbox{E}\kern-.125emX}}

\usepackage[colorlinks,linkcolor=black,anchorcolor=black,citecolor=black]{hyperref}
\usepackage{multirow}
\usepackage{multicol}
\usepackage{lineno}
\usepackage{amsmath}
\usepackage{algorithmic}
\usepackage{booktabs}
\usepackage{caption}
\usepackage[inline]{enumitem}
\usepackage{algorithm}
\usepackage[algo2e]{algorithm2e} 
\usepackage{times}
\usepackage{epsfig}
\usepackage{amssymb}
\usepackage{subcaption}
\usepackage{amsfonts}
\usepackage{array}
\usepackage{textcomp}
\usepackage{makecell}
\usepackage{diagbox}
\usepackage{threeparttable}
\usepackage{mathtools}
\usepackage{eqparbox}
\usepackage{bm}
\usepackage{bbding}
\usepackage{comment}
\usepackage{color}
\usepackage{bbding}
\usepackage{pifont}
\usepackage{wasysym}
\usepackage{stfloats}
\usepackage{float}

\usepackage[left=1.7cm,right=1.7cm,top=1.7cm,bottom=1.7cm]{geometry}
\renewcommand{\baselinestretch}{0.97}  % 0.95倍行距

\begin{document}

\title{SpecRef: A Fast Training-free Baseline of Specific Reference-Condition Real Image Editing
\thanks{$^{*}$ Corresponding author. \textsuperscript{$\dagger$} Equal contributions. This research was funded by Shenzhen Science and Technology Program (JCYJ20220818101607015).
}
}

\author{\IEEEauthorblockN{Songyan Chen\textsuperscript{1,2,$\dagger$}, Jiancheng Huang\textsuperscript{1,2,$\dagger$,$*$}}
\IEEEauthorblockA{\textsuperscript{1}\textit{Shenzhen Institute of Advanced Technology, Chinese Academy of Sciences} \\
\textsuperscript{2}\textit{University of Chinese Academy of Sciences} \\
$^{*}$ Corresponding author: \textit{jc.huang@siat.ac.cn}}}

\maketitle

\begin{abstract}
Text-conditional image editing based on large diffusion generative model has attracted the attention of both the industry and the research community. 
Most existing methods are non-reference editing, with the user only able to provide a source image and text prompt. However, it restricts user's control over the characteristics of editing outcome. To increase user freedom, we propose a new task called Specific Reference Condition Real Image Editing, which allows user to provide a reference image to further control the outcome, such as replacing an object with a particular one. To accomplish this, we propose a fast baseline method named SpecRef. Specifically, we design a Specific Reference Attention Controller to incorporate features from the reference image, and adopt a mask mechanism to prevent interference between editing and non-editing regions. We evaluate SpecRef on typical editing tasks and show that it can achieve satisfactory performance. The source code is available on \url{https://github.com/jingjiqinggong/specp2p}.
\end{abstract}


\section{Introduction}\label{sec:intro}

%生成大模型火热的背景
The diffusion model~\cite{ho2020denoising,song2019generative,nichol2021improved}, as a new large image generative model, has revolutionised the field of image generation and made the tasks related to AI text-to-image generation~\cite{shi2023instantbooth,ruiz2022dreambooth,gal2022image,ho2022cascaded} very hot. 
Since the limitations of the traditional image editing task~\cite{10268075,liu2023deseal} are very obvious and its performance is unsatisfactory~\cite{richardson2020encoding,zhu2020improved,crowson2022vqgan}, researchers start to utilize the large image generative model to perform image editing. Based on AI text-to-image generation, text-conditional image editing~\cite{nichol2021glide,meng2021sdedit,huang2023kv,chen2023fec,huang2023seal2real} is a very new and interesting task that has emerged in recent months with great commercial value and immeasurable potential. 

%介绍SD
Among these diffusion models~\cite{avrahami2022blended,song2020denoising,kim2022diffusionclip,huang2023wavedm,huang2023bootstrap}, the Stable Diffusion Model~\cite{rombach2021highresolution} has emerged as a formidable contender. Stable Diffusion model adapts diffusion processes in latent space rather than the original RGB space. Besides, it use cross-attention layer for the input of text prompt, yielding high-fidelity results characterized by remarkable visual coherence and semantic fidelity.
Thanks to its powerful text-conditional image generation capabilities, most current image editing methods are based on Stable Diffusion Model~\cite{rombach2021highresolution}. 

%引出基于SD的图像编辑这个领域
However, due to the lack of available large paired datasets, many methods tend to use training-free designs~\cite{hertz2022prompt,brooks2022instructpix2pix,cao2023masactrl}. Such designs involve exploring some of the internal features or structures of the pre-trained stable diffusion model itself, analysing the usefulness of these internal features and structures, and artificially manipulating them to achieve image editing results.

\begin{figure}[t]
\centering
\includegraphics[width=1.0\linewidth]{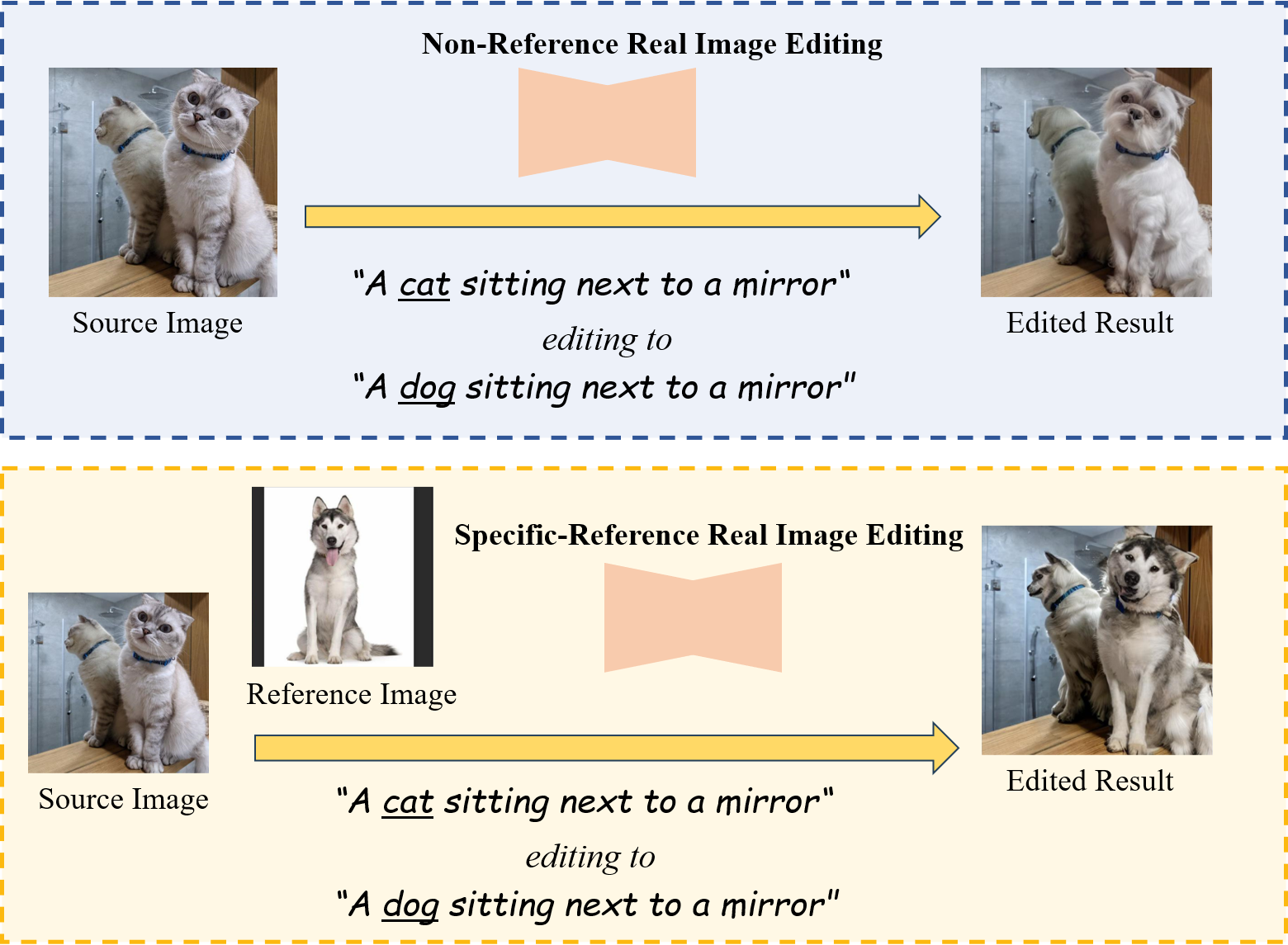}
\caption{A demonstration of the existing non-reference editing task and our new task. The top row is the existing task and the bottom row is our specific reference condition image editing task.}
\label{fig:task}
\end{figure}  

%介绍baseline方法P2P
Among them, the well-known image editing work P2P~\cite{hertz2022prompt} tapped into a prior of the pre-training Stable Diffusion, which found that the cross-attention map corresponding to each word in the text prompt in the cross-attention layer corresponded spatially to the objects in the generated image. For example, given a text prompt "a photo of a cat", the cross attention map of "cat" will overlap with the cat area. With this prior, P2P generates consistency images by controlling the cross-attention map. However, P2P is proposed for synthetic image editing instead of real image editing. For adapting P2P to real image, Null-Text Inversion~\cite{mokady2022null} is proposed to invert a real image to noisy latent space. Then, P2P+Null-Text becomes a common real image editing baseline, which can perform object replacement task.

%提出P2P的问题
Although P2P can perform object replacement well, the user cannot decide the target result. For example, as in Fig.~\ref{fig:edit1} p2p(no img2), P2P can replace the cat with a dog, but the user can't decide what the dog looks like. If the user wants to replace that cat with a desired dog, P2P has no way of achieving this goal. Thus, we define all these kinds of editing methods as non-reference editing, while we propose a new kind of task named specific reference-condition editing. 

%简单介绍我们的任务
Here we introduce our specific reference editing task. Just like non-reference editing, in specific reference condition editing, the user can also provide a source image, a source prompt, and a target prompt. In addition, our task can provide an additional input image as a reference image, which is the main difference with non-reference editing. Then, our goal is to edit the source image according to the target prompt and the reference image.

%简单介绍我们的方法的思路来源
In order to propose a baseline for this new task, we first need to find a reasonable way to feed the reference information into the network. As we all know, Stable Diffusion (SD)~\cite{rombach2022high} has the U-Net~\cite{ronneberger2015unet} as the noise predict network $\epsilon_\theta$ for predicting noise. There are several phenomenons~\cite{hertz2022prompt,tumanyan2022plug,chefer2023attend} in the attention layers of $\epsilon_\theta$: 1) The attention map of cross-attention determines the structure of the image (the position of the appearing objects). 2) $K$ and $V$ of the cross-attention layers change the texture and detail of generated image. 3) $K$, $V$ and the attention map in self-attention layers have a dramatic effect on the generated result in terms of content.

Based on the above phenomena, we logically think that we can use the $K$ and $V$ in the self-attention layers of the reference image as reference features to represent the content of the reference image. Then we can use these reference features in the editing stage to bring in the content from the reference image.
%三个贡献点
Our main contributions are summarized as follows. 
\begin{itemize}
\setlength{\itemsep}{0pt}
\setlength{\parsep}{0pt}
\setlength{\parskip}{0pt}
\item We analyse the drawback of non-reference editing and describe our new task specific reference editing to enhance user freedom.
\item We propose a fast and training-free baseline method for this task named SpecRef, which can change the edited part of the source image according to the reference image.
\item Comprehensive Experiments show that SpecRef as a baseline method can achieve satisfactory editing performance in the specific reference-condition real image editing, demonstrating the superiority of this new task over previous non-reference editing qualitatively.
\end{itemize}

 \begin{figure*}[t]
\centering
\includegraphics[width=1.0\linewidth]{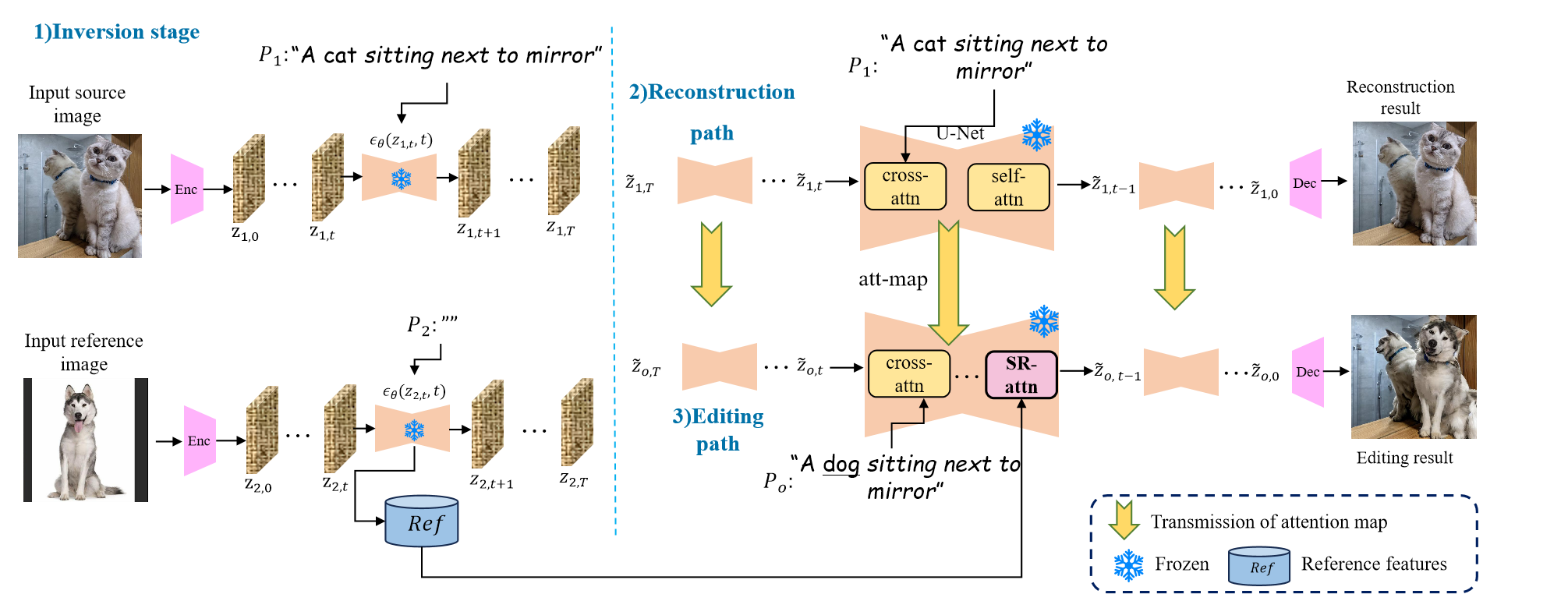}
\caption{The pipeline of SpecRef, consisting of two stages, inversion stage and editing stage. During inversion stage, we perform inversion on both source image $I_1$ and reference image $I_2$ to obtain the noisy latents and reference features. Then in the editing stage, there are two paths, reconstruction path for reconstruct $I_1$ and editing path for generating the edited result.}
\label{fig:pipe}
\end{figure*}  

\section{Method}

We first define our new task specific reference-condition image editing in Sec.~\ref{sec:task}. Then we introduce the motivation of this task and setting in Sec.~\ref{sec:mot}. Finally, we analyse this task and propose our first baseline method for this task in Sec~\ref{sec:specref}.

\subsection{Introduction of Our Task}\label{sec:task}

In this section, we define our new proposed task. At the beginning, given a input image $I_1$ as source image and a input image $I_2$ as reference image, and we also provide two text prompt $P_1, P_2$ as source prompt and a text prompt $P_o$ as target prompt. 

The goal of our new task is to generate a edited result image $I_o$ with a pre-trained Stable Diffusion Model. This edited result image $I_o$ should be consistent with the source image $I_1$ in the unedited part and be similar to the reference image $I_2$.

For example, as shown in Fig.~\ref{fig:task}, given a source image $I_1$ with text prompt $P_1$ ``a cat sitting next to a mirror", and then we edit the text prompt $P_o$ to ``a tiger sitting next to a mirror" and provide a reference image $I_2$ with a tiger and its prompt $P_2$. Our task is to generate a new image with the given tiger sitting next to a mirror, while the other contents of the image remain basically unchanged.

The task is in fact very difficult and, according to our full investigation, no current work is focusing on this task, since most of them are designed non-reference image editing~\cite{hertz2022prompt}.

\subsection{Motivation of Our Task}\label{sec:mot}

We now discuss in detail why this task needs to be proposed. First of all, existing image editing tasks can only provide a real image as input, which is equivalent to only input $I_1$ in our task. There is a lot of current work~\cite{crowson2022vqgan,kawar2022imagic,hertz2022prompt,brooks2022instructpix2pix} focusing on this setting, which is called non-reference editing in this paper. 

However, this non-reference editing is too limited for users to satisfy their diverse needs. For example, as shown in Fig~\ref{fig:edit1}, the user wants to edit the cat in the first row of Fig~\ref{fig:edit1} into a tiger. However, the user cannot manipulate the specific appearance of the generated tiger by non-reference editing. For instance, as in the p2p(no img2) of Fig~\ref{fig:edit1}, we show the results of a common non-reference editing method prompt-to-prompt~\cite{hertz2022prompt}(abbreviate as P2P). If the user wants to decide what the generated tiger looks like, our task is exactly the right solution to their needs.

Besides, as shown in the fourth row of Fig~\ref{fig:edit1}, it is difficult to describe the content of the desired dress as ``colorful plaid dress". Thus, in the p2p(no img2) of Fig~\ref{fig:edit1}, non-reference editing task can not generate a ideal colorful dress.

\subsection{Implementation Motivation}

As mentioned in Sec.~\ref{sec:intro}, features in the self-attention layer (such as Key and Value) have a strong influence on the content and texture of the generated object. Therefore, if we need to accomplish our new task of introducing the content of the reference image $I_2$ into the model, then utilising features in the self-attention layer is a very good choose. 

Based on~\cite{cao2023masactrl}, we know that if we can extract the features of $I_2$ in the self-attention layer and use them in the generation process, then we can reasonably generate content similar to $I_2$. Next, we describe how to use the features of $I_2$ in the self-attention layer, which is called as reference features in this paper.

\begin{algorithm}[t]
\SetAlgoLined
\textbf{Require:} the latent of the original real image $z_{1,0}$ and reference image $z_{2,0}$, the source prompt $P_1$ ($P2=``")$, the target prompt $P_o$ and the mask\_s $M_s$.
 \vspace{1mm} \hrule \vspace{1mm}
\textbf{1) Inversion stage:} Utilize DDIM inversion Eq.\ref{invert} to obtain the reference features(self attention KV) $Ref$ of image $I_2$ and intermediate latents $\{z_{1,t}, t=1,...,T \}$ of source image $I_1$. 
 \vspace{1mm} \hrule \vspace{1mm}
 \textbf{2) Editing stage:} 
 Set the beginning latent $\tilde{z}_{o,T} = z_{1,T}$.\\
 \For{$t=T,T-1,\ldots,1$}{
       Calculate mixture feature of Specific Reference Attention Layer $\epsilon_\theta$ as Eq.\ref{eq:attention},\ref{eq:mask} and obtain the latent noise for editing path:\\
        $\epsilon_t=\epsilon_\theta(\tilde z_{o,t},\mathbf{c}_o,t;Ref,M_s,l_t)$;\\
        $\tilde z_{o,t-1} = \mathrm{PrevStep}(\tilde z_{o,t}, \epsilon_t)$ as Eq.~\ref{sampling};\\
 }
 $I_o = \mathrm{Decode}(\tilde{z}_{o,0})$;\\
 \textbf{Return} Editing result $I_o$
\caption{The 2 Stages of SpecRef}\label{alg}
\end{algorithm}
\subsection{Implement of Our SpecRef}\label{sec:specref}

\subsubsection{Extraction of Reference Feature}
\text{}

As illustrated in the inversion stage of Fig.~\ref{fig:pipe}, for extracting the Key $K$ and Value $V$ features in the self-attention layer of $I_2$, we first perform inversion of $I_2$ to get a series of latent noises $\{z_{2,t}\}$. 
During this inversion for $I_2$, we save the Key $K$ and Value $V$ features in the self-attention layer of $\epsilon_\theta(z_{2,t},t)$ at each step $t$.

% Here, we define the inversion formula for $I_2$:
\begin{equation}\label{invert}
\begin{aligned}
    z_{2,t} = & \sqrt{\alpha_t}  \frac{z_{2,t-1} - \sqrt{1-\alpha_{t-1}} \cdot \bar\epsilon_\theta(z_{2,t-1},t-1) }{\sqrt {\alpha_{t-1}}} \\ &+ \sqrt{1-\alpha_{t}} \cdot \bar\epsilon_\theta(z_{2,t-1},t-1)
\end{aligned}
\end{equation}
where $\bar\epsilon_\theta(z_{2,t-1},t)$ means that we preserve the $K$ (Key) and $V$ (Value) of the self-attention layer in the network of $\epsilon_\theta$. We denote $\bar{KV}_t = \{\bar{KV}_{t,1}, ... ... , \bar{KV}_{t,L}\}$, where $\bar{KV}_{t,l} = (\bar{K}_{t,l}, \bar{V}_{t,l})$, and $l$ represent the $l$-th Transformer block. The preservation process is the only difference between it and DDIM inversion. Note that $Ref = \{\bar{KV}_1, ..., \bar{KV}_T\}$ is called \textbf{reference features} in this paper because they represent the content of the reference image $I_2$.

\subsubsection{Specific Reference Attention Layer}
\text{}

After obtaining the above reference features, we also perform inversion of $I_1$ to get a series of latent noises $\{z_{1,t}\}$ including $z_{1,T}$. $z_{1,T}$ can be used as the beginning of the editing stage.
\begin{figure}[t]
\centering
\includegraphics[width=1.0\linewidth]{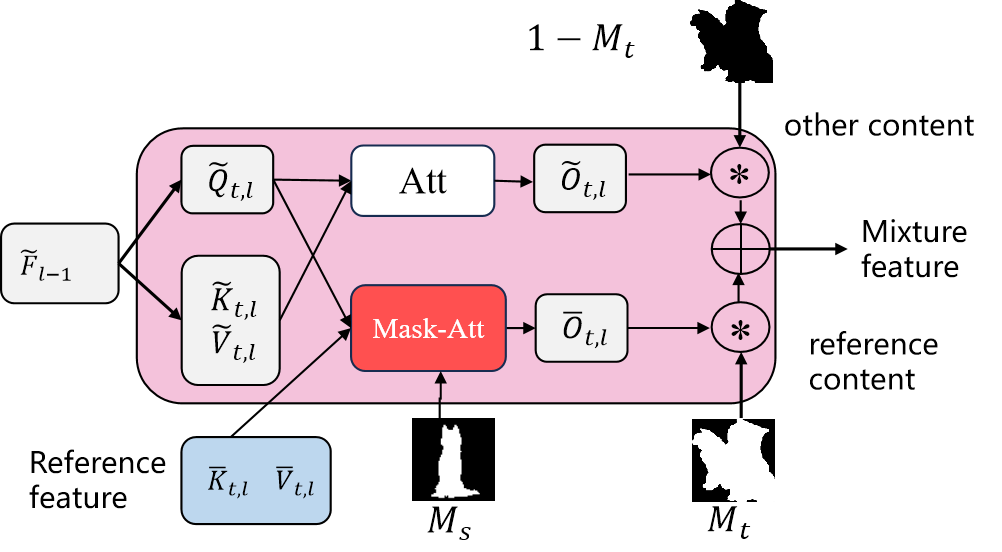}
\caption{The proposed Specific Reference Attention Layer (SR-attn).}
\label{fig:SR-attn}
\end{figure} 

\begin{figure*}[t]
\centering
\includegraphics[width=1.0\linewidth]{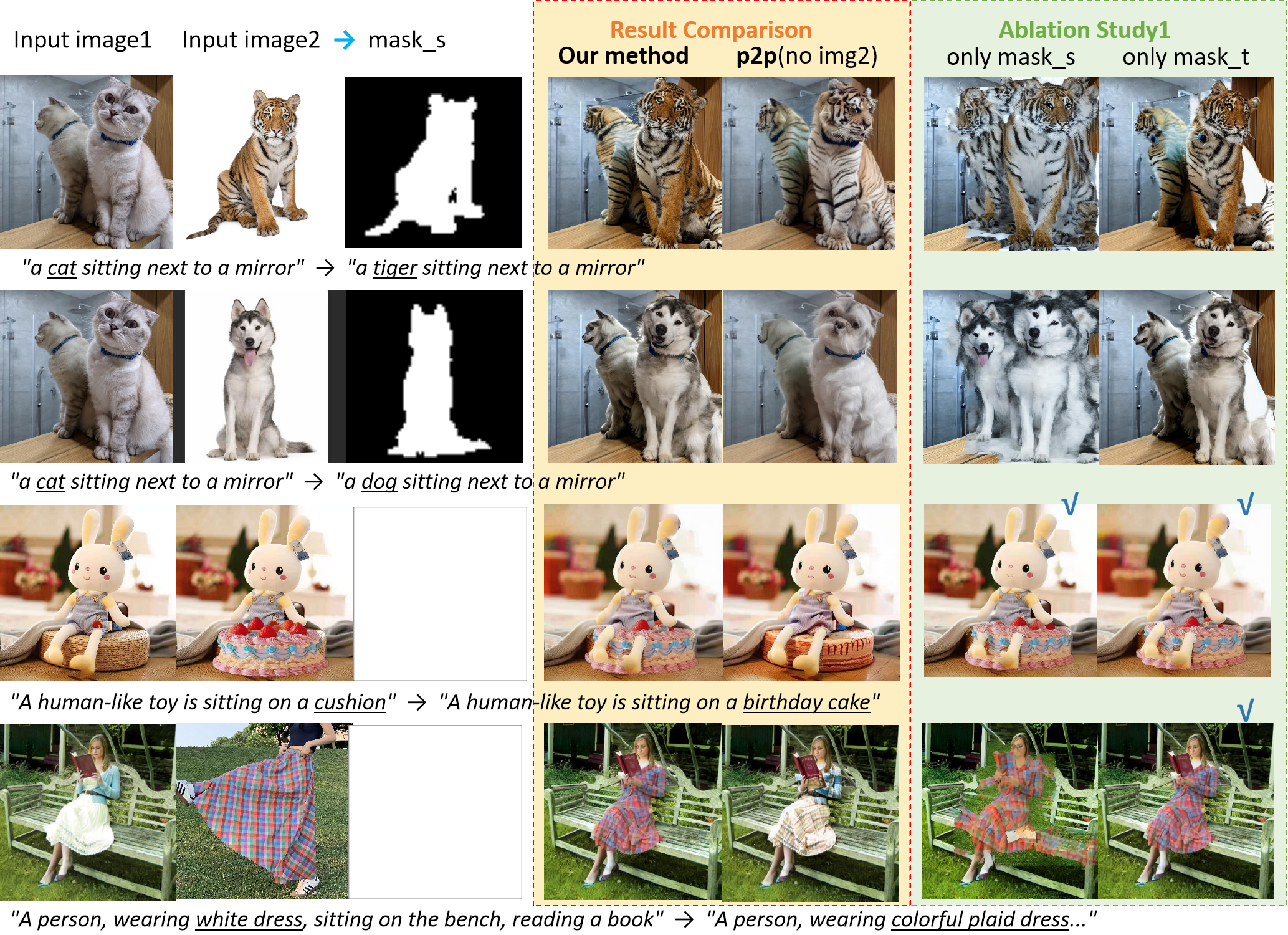}
\caption{The experimental results. Our SpecRef can solve the problem whereby non-reference editing (p2p) fails to generate content for certain words, and replace the object by specific reference. }
\label{fig:edit1}
\end{figure*}

Based on the structure of P2P~\cite{hertz2022prompt}, our SpecRef also has two path during editing stage. The first path is the reconstruction path with the noisy latents $\tilde z_{1,t}$ and the other path is the editing path with the noisy latent $\tilde z_{o,t}$ . As mentioned in Sec.~\ref{sec:intro}, in P2P, editing path is for generating the target image $I_o$, while the information of $I_1$ is transmitted from the reconstruction path. However, P2P can only edit by text without providing the specific reference image. For instance, in second row of Fig.~\ref{fig:edit1}, p2p(no img2) can only edit the cat to be a dog, but can not control the specific dog.

For providing the specific reference image $I_2$, we need to modify the attention layer in network $\epsilon_\theta(\tilde z_{o,t},P_o,t;Ref,l_t)$ during sampling stage, where $l_t$ denotes the number and place to insert the new layer. Now we define our Specific Reference Attention layer (SR-attn) as shown in Fig~\ref{fig:SR-attn}. We use $M_s^{inf}$ to denote the result of setting zero element in $M_s$ to negative infinity. Given the $l$-th self-attention layer during timestep $t$ of reverse process in $\epsilon_\theta(\tilde z_{o,t},P_o,t;Ref,l_t)$, we have:
\begin{equation}
    \label{eq:attention}
    \begin{aligned}
    %     \text{Attn}(\tilde{Q}_{t,l}, \bar{K}_{t,l}) = \bar A_{t,l}= \text{Softmax}(\frac{\tilde{Q}_{t,l}{\bar{K}_{t,l}}^T}{\sqrt{d}}),\\
    % \text{Attn}(\tilde{Q}_{t,l}, \tilde{K}_{t,l}) = \tilde A_{t,l}= \text{Softmax}(\frac{\tilde{Q}_{t,l}{\tilde{K}_{t,l}}^T}{\sqrt{d}}),
    \text{MaskAttn}(\tilde{Q}_{t,l}, \bar{K}_{t,l},M_s) = \text{Softmax}(\frac{\tilde{Q}_{t,l}{\bar{K}_{t,l}}^T}{\sqrt{d}} + M_s^{inf}),
    \end{aligned}
\end{equation}
where $\tilde{Q}_{t,l}$ and $\tilde{K}_{t,l}$ denote the Query and Key features in the self-attention layer of $\epsilon_\theta(\tilde z_{o,t},t;Ref,l_t)$. $\bar{K}_{t,l}$ and $\bar{V}_{t,l}$ represent the reference features in $Ref$. We don't want to compute attention between the whole reference features from $I_2$ and the whole target feature $\tilde{Q}_{t,l}$, which will lead to regional chaos as shown in Fig.~\ref{fig:edit1} ablation study1.

Thus, the ideal attention should be computed between the reference object in $I_2$ (such as the tiger in first row of Fig.~\ref{fig:edit1}) and the edited part of $I_1$ (such as the cat in first row of Fig.~\ref{fig:edit1}). We realise this by using attention mask in the SR-attn. Here we use two types of mask. The source mask
$M_s$ denotes the reference object in $I_2$ as the mask\_s in Fig.~\ref{fig:edit1}. We use the cross-attention map in P2P~\cite{hertz2022prompt} as the target mask $M_t$.
\begin{equation}
    \label{eq:mask}
    \begin{aligned}
        &\bar O_{t,l} =\text{MaskAttn}(\tilde{Q}_{t,l}, \bar{K}_{t,l},M_s)\bar{V}_{t,l},\\
        &O_{t,l} = M_t \bar O_{t-1} + (1-M_t) \text{Attn}(\tilde{Q}_{t,l}, \tilde{K}_{t,l})\tilde{V}_{t,l}
    \end{aligned}
\end{equation}
where $\text{MaskAttn}()$ means that only keep the attention of the $M_s$ region and mask the other regions with zero (no attention). We can see that the content of editing part $M_t$ is from $\bar{V}_{t,l}$ and non-editing part $1-M_t$ is from $\tilde{V}_{t,l}$. With this SR-attn layer, we define our editing sampling step in the editing stage as:
\begin{equation}\label{sampling}
\begin{aligned}
    \tilde z_{o,t-1} = &
\sqrt {\alpha_{t-1}} \frac{\tilde z_{o,t} - \sqrt{1-\alpha_t} \cdot \epsilon_\theta(\tilde z_{o,t},P_o,t;Ref,l_t,Mask_s) }{\sqrt {\alpha_t}}
\\ & + \sqrt{1-\alpha_{t-1}} \cdot \epsilon_\theta(\tilde z_{o,t},P_o,t; Ref,l_t,Mask_s),
\end{aligned}
\end{equation}
where $\epsilon_\theta(\tilde z_{o,t},P_o,t; Ref,l_t)$ denotes the noise estimation network using SR-attn layer within the range of $l_t=(l_{t,s},l_{t,e})$. After each sampling step during concurrent reconstruction and editing process, we follow P2P to use its local blending operation as:
\begin{equation}\label{p2p}
\begin{aligned}
    \tilde z_{o,t-1} = \mathrm{Blend}(\tilde z_{o,t-1}, \tilde z_{1,t-1}),
\end{aligned}
\end{equation}
Fig.~\ref{fig:pipe} and \ref{fig:SR-attn} show that this process. The algorithm of our SpecRef with inversion and sampling stage is provided in Algorithm~\ref{alg}. With the proposed SR-attn layer, we rationalise the information of the reference features $Ref$ into the network $\epsilon_\theta(\tilde z_{o,t},P_o,t; Ref,l_t)$ during the editing stage within certain time-step ranges, making the editing results fully controllable.

\section{Experiment}

\begin{figure*}[t]
\centering
\includegraphics[width=1.0\linewidth]{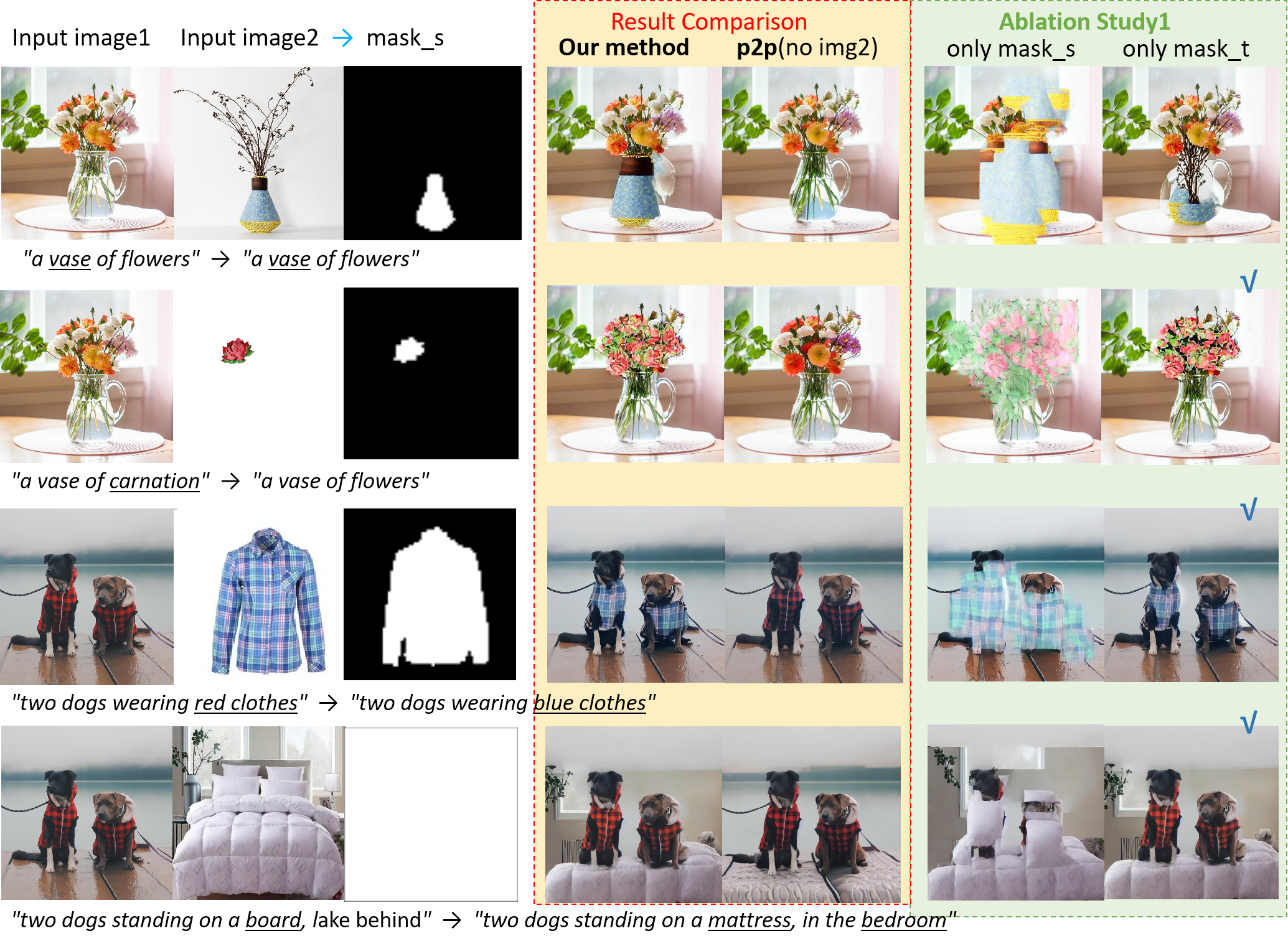}
\caption{The experimental results. Our SpecRef can solve the problem whereby non-reference editing (p2p) fails to generate content for certain words, and replace the object by specific reference. }
\label{fig:edit2}
\end{figure*} 

In the experiments for our task, the inputs include the source image $I_1$, the reference image $I_2$, and the mask\_s image $M_s$. The output is the edited result image. We perform common editing tasks, including object replacement, clothing replacement, and scene replacement. Since what we proposed is a new task, we only perform a visual comparison with the common non-reference editing method P2P~\cite{hertz2022prompt}.

\subsection{Comparison with Non-Reference Editing}
% since P2P is not a method for real image editing, we use Null-text inversion~\cite{mokady2022null} to make P2P work on real images.
% make up for the defects of uneditable words in P2P
% solve the problem whereby non-reference editing (p2p) fails to generate content for certain words
We compare our task with naive non-reference editing method P2P to show the effect of the reference image on editing. We adapt P2P\cite{hertz2022prompt} to real imgae editing by using FEC-ref\cite{chen2023fec}, where each reconstruction step uses the corresponding latent from inversion as input to UNET. As the results of Result comparison in Fig.~\ref{fig:edit1} and Fig.\ref{fig:edit2}, there are two obvious advantages of our task: 1) Our task can solve the problem whereby non-reference editing (p2p) fails to generate content for certain words. For example, P2P cannot change the color of dog's clothes in third row of Fig.~\ref{fig:edit2}. 2) Our task can replace object by specific reference. In Fig.~\ref{fig:edit1}, the target tiger, dog, dress and cake all look the way we specify. The reason why our task can be successful is that we use the reference features in the editing stage.

\begin{figure}[hbtp]
\centering
\includegraphics[width=0.9\linewidth]{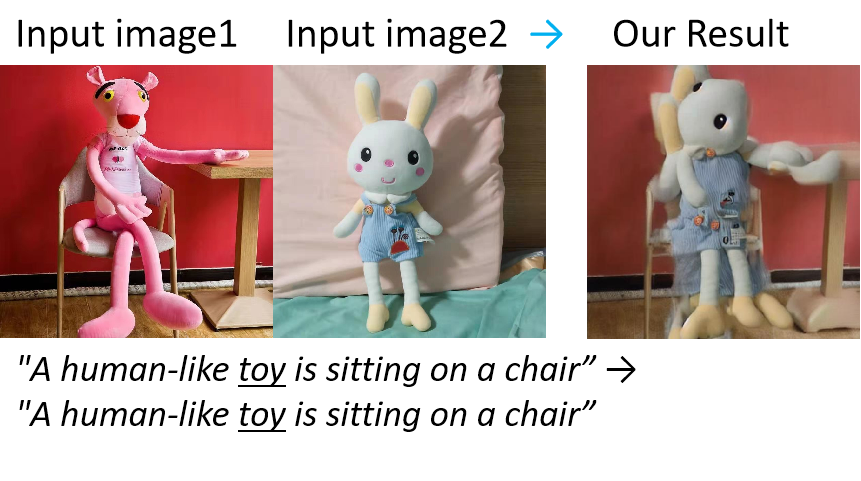}
\caption{Examples of editing failures}
\label{fig:fail}
\end{figure}  

\subsection{Ablation Study}
We now discuss the implications of the reference image and the two masks used in the proposed Specific Reference Attention Layer. First, we review the roles of the two masks. mask\_s $M_s$ is used to cut out the object in the reference image, which can either come from user or be automatically segmented using the segment anything model (SAM)~\cite{kirillov2023segment}. $M_s$ has the same for each step. mask\_t $M_t$ is used to represent the editing region during sampling process. According to the editing word, we use the cross attention map~\cite{hertz2022prompt} during editing to automatically get the mask of the editing region. $M_t$ is different at each step. Reasonable use of $M_t$ and $M_s$ makes the attention of the editing region focus on the reference image, while the attention of the non-editing region focuses on the editing image itself.

As we can see from the resulting image, mask\_t is important. If you look at the only mask\_s column. Without mask\_t, the reference item modifies the rest of it as well. This leads to the appearance of repeated textures, interference with colour textures, white edges in the editing area, etc.
The reason for this failure is because the lack of mask\_t affects the extent of the edit area behind this step, causing the edit area itself to change incorrectly.
As we can see from the resulting images, mask\_s can do some lifting. For example, in the result presentation, we use different types of reference images, e.g., some reference images have only objects and some reference images also contain backgrounds. Looking at the image only mask\_t column, we can barely get acceptable results in the absence of mask\_t. We conclude that the inspiration is: you can key out the objects in the reference image and paste them into the original image, so that you can get a more natural effect.

\subsection{Limitations}\label{sec:lim}
Of course, as a baseline method, we also have more disadvantages and limitations. Firstly, it is easy to fail when the area where the reference image is located is very far from the editing area of the original image. As Fig.~\ref{fig:fail} shows, this can lead to obvious duplication of regions, colours or textures in the edited result. For example, in the first row of Fig.~\ref{fig:fail}, the rabbit in the edited result has a strange head and too many legs. This is due to the fact that Specific Reference Attention Layer is a cross-attention based operation, as in Equation \ref{eq:attention}, where the content from the reference image $I_2$ is passed through the cross-attention to the editing result image $I_o$. However, this operation is based on the assumption that the region of the reference target and the edited region of the original image need to be similar in size and shape. When a sample image is encountered that does not meet this assumption, it is easy to edit it to an unnatural result. This is a problem that needs to be solved in this field as soon as possible.

\section{Conclusion and Future Work}
In this paper, we first analyse the limitation of the existing task non-reference image editing and describe why we need a new task for users. Then, we define our new task specific reference editing. After that, we propose a fast and training-free baseline method SpecRef for this task, which can change the edited part of the source image according to the reference image. We conduct comprehensive experiments to show that SpecRef can achieve satisfactory performance in specific reference-condition real image editing task.

However, as in Sec.~\ref{sec:lim}, SpecRef also has many limitations and can not be robust in all scenes. Therefore, we analyse the reasons of the fail cases and struggle to design a training-based robust methods in the future.

\bibliographystyle{IEEEtran}
\bibliography{egbib.bib}

\end{document}